\documentclass[10pt, conference, compsocconf]{IEEEtran}

\ifCLASSINFOpdf

\else

\fi

\usepackage{amssymb}
\usepackage{graphicx}
\usepackage{times}
\usepackage{epsfig}
\usepackage{amsmath}
\usepackage{subfigure}
\usepackage{amssymb}
\usepackage{epstopdf}
\usepackage{url}
\usepackage{paralist}
\usepackage{multicol}

\usepackage{url}

\begin{document}
%
\title{An Immersive Telepresence System using RGB-D Sensors and Head Mounted Display}

\author{
    \IEEEauthorblockN{Xinzhong Lu\IEEEauthorrefmark{1}, Ju Shen\IEEEauthorrefmark{1}, Saverio Perugini\IEEEauthorrefmark{1}, Jianjun Yang\IEEEauthorrefmark{2}}
    \IEEEauthorblockA{\IEEEauthorrefmark{1}Department of Computer Science, University of Dayton}
    \IEEEauthorblockA{\IEEEauthorrefmark{2}Department of Computer Science, University of North Georgia}
}

\maketitle

\begin{abstract} We present a tele-immersive system that enables people to
interact with each other in a virtual world using body gestures in addition to
verbal communication. Beyond the obvious applications, including general online
conversations and gaming, we hypothesize that our proposed system would be
particularly beneficial to education by offering rich visual contents and
interactivity. One distinct feature is the integration of egocentric pose
recognition that allows participants to use their gestures to demonstrate and
manipulate virtual objects simultaneously. This functionality enables the
instructor to effectively and efficiently explain and illustrate complex
concepts or sophisticated problems in an intuitive manner.  The highly
interactive and flexible environment can capture and sustain more student
attention than the traditional classroom setting and, thus, delivers a
compelling experience to the students.  Our main focus here is to investigate
possible solutions for the system design and implementation and devise
strategies for fast, efficient computation suitable for visual data processing
and network transmission. We describe the technique and experiments in details
and provide quantitative performance results, demonstrating our system can be
run comfortably and reliably for different application scenarios. Our
preliminary results are promising and demonstrate the potential for more
compelling directions in cyberlearning.  \end{abstract}


\IEEEpeerreviewmaketitle

\section{Introduction}
With the advent of internet and multimedia technology, online video chatting
(e.g., Skype) has become a popular communication tool in daily life. The
face-to-face visual effects and synchronized voices allow people to talk
remotely with convenience. The advantages of reduced time and financial cost,
and higher flexibility, make it widely used in many tasks, such as business
conferences and job interviews. However, despite the benefits for general
communication purposes, online video chatting has inherent difficulties to be
adapted for more specific uses. For example, in the process of training, or
classroom teaching, more interactivity is necessary.  For example, the teacher
may use an object or a tool to demonstrate and explain some sophisticated
concepts.  Body movement or gesture sometimes can offer important cues to
improve and enhance the absorption, understanding, and retention of the
material.  For such scenarios, the traditional online video chatting may fail
to perform desirably. The typical fixed webcam setting with static perspective,
narrow field of view, and limited interactivity has limited capabilities.  To
mitigate the limitations, virtual reality (\textsc{vr}) can be an appropriate
compensation that utilizes advanced vision, graphics techniques, and novel
hardware to create a more friendly and immersive experience.
\begin{figure}[!htb]
\centerline{\epsfig{figure=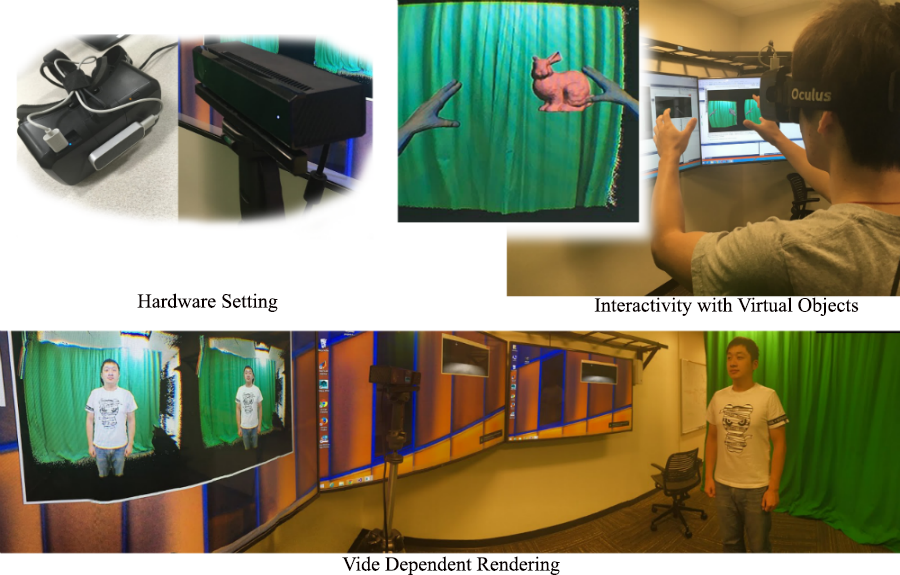, width=9cm}}
\caption{The overview of our proposed Tele-immersive system.}
\label{fig:overview}
\end{figure}

Recently, tele-immersive (\textsc{ti})-based systems are gaining popularity in
both research and practice, aiming to provide a natural and immersive user
experience with more intuitive interactions and better visual quality.  Some
commercial conferencing systems, such as Cisco Telepresence and Hewlett-Packard
Halo systems, demonstrate exciting features, including life-size video view,
stereo audio, and high visual effects of rendering.  Compared to the
traditional webcam-based video chatting, these systems represent a substantial
improvement.

Despite these improvements, these systems still fall short of
offering realistic immersive experiences.  One common and critical defect is
the lack of the view dependent rendering. The observed visual contents from the
screen are relatively static against the user's view point, which is not
consistent with real-world perception. This absence of this feature fails to effectively preserve mutual eye gaze and
potentially distracts users' attention.
Another drawback of existing systems is the exclusive use of conventional
interaction, which only concentrates on oral communication and ignores
non-verbal cues such as body gestures. For many online interactions, or
collaborations, non-verbal communication has the potential expressivity to
vividly convey useful information. Thus, a desirable tele-immersive system
should satisfy the following properties:

\begin{itemize}

\item \textbf{Correct co-location geometry between users}:
during telecommunication,
users are projected into a particular virtual environment. To offer the sense
of face-to-face communication, we need to accurately place each user in the
right coordinate space to provide the expected virtual perception.

\item \textbf{Dynamic view dependent rendering}: to generate realistic visual
effects and enhance eye gaze, we need to render the screen based on the user's
viewpoint position in the virtual space. As the viewpoint moves, the observed
objects should change accordingly so that they conform to the perception
principle in physical world.

\item \textbf{Rich non-verbal interaction}: by applying \textsc{vr} technology,
the platform is empowered with the capability of capturing non-verbal cues to
enrich the interactivity. Information such as body gesture, physical distance,
and size, should be incorporated with audiovisual media for better accuracy,
higher flexibility, and efficiency.

\item \textbf{Realtime performance}: as an essential prerequisite, the system
should run at interactive speed to offer liberal communication in unobstructed
space. For a typical \textsc{ti} system, most of the computational load
concentrates on two components: visual content generation and network
transmission. It is crucial to make the system computationally tractable for
automatic recognition, fast rendering, and large visual data transportation
through optimized strategies.

\end{itemize}

Motivated by these requirements, we present a highly interactive,
tele-immersive system for effective remote communication. We anticipate the
proposed technique will offer a flexible platform for a wide range of
applications, especially for educational purposes and applications.  The system
setup is shown in Fig.~\ref{fig:overview}.  We use head-mounted display
(\textsc{hmd}) as the communication media instead of traditional desktop-size
displays; a pair of low-cost \textit{Kinect} (\textsc{rgb-d}) cameras and
standard microphones used to capture user's input; an infrared light sensor is
utilized for gesture capture and recognition from egocentric perspective (i.e.,
first-person view).

Our goal is to enable users from remote places to communicate and collaborate
in a realistic face-to-face manner. Two distinct features of our system are:
\begin{inparaenum}[i)] \item realtime view-dependent rendering is offered so
that the displayed content updates dynamically according to the view position
and orientation, and \item users are allowed to directly operate on virtual
objects in an intuitive manner.  As the example in Fig.~\ref{fig:overview}
illustrates, the teacher uses the bunny object to explain a particular concept.
He can hold, rotate, and manipulate it with her hands. Moreover, multi-users
can work collaboratively manipulate the shared virtual objects in a
synchronized manner.  Each user can not only see each other, but also observe
the consequences of each other's actions reflected on the virtual objects. For
instance, if both users push a virtual ball from different directions, they
will perceive the transformation (i.e., shape or position changes) of the ball
caused by their actions.
\end{inparaenum}

However, achieving these desired functionalities is challenging.  First, to
accurately generate realistic perspective views, we need to acquire correct 3D
scene geometry and user's viewpoint positions. Second, user's body poses need
to be detected and tracked in realtime. Third, to effectively produce user's
interactions among each other and with the surrounding virtual environment, the
system should be able to handle coordinate synchronization, collision detection
by incorporating physical properties, such as object contours, and weight.
Thus, our primary efforts are directed toward designing, implementing, and
optimizing the system by identifying each module individually and collectively
integrating them.

\section{System Architecture}

The overview of our proposed \textsc{ti} system is depicted in
Fig.~\ref{fig:overview} with the hardware setup and user demonstration. At each
user's end of the proposed client-to-client scenario, an \textsc{rgb-d}-based
system, a head-mounted display (\textit{Oculus Rift}), and motion detection
sensor operate together for input capture and data processing. For the
\textsc{rgb-d} acquisition, we choose the \textit{Kinect} camera as it is
widely accessible and low-cost. For the motion detection, the
\textit{LeapMotion} device is used, which is a portable device consisting of
two cameras and three infrared \textsc{led}s. Since the novelty of our system
is designed primarily for producing rich visual effects, we only focus on the
video-sensing devices in our prototype. 

\subsection{Functionalities}

\begin{figure}[!htb]
\centerline{\epsfig{figure=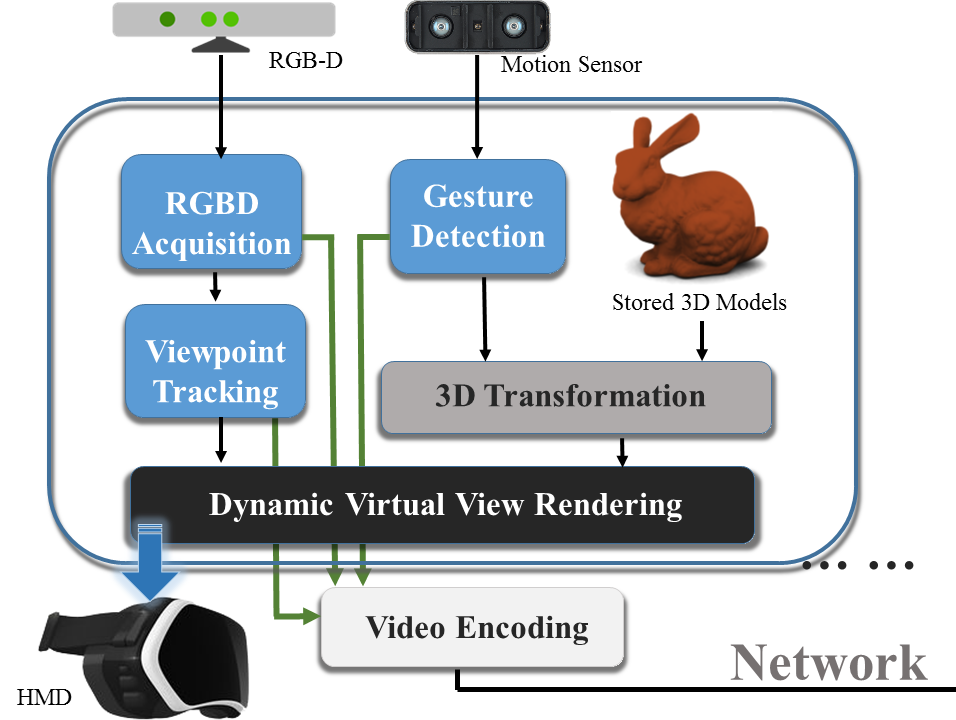, width=8cm}}
\caption{System architecture and pipeline.}
\label{fig:pipeline}
\end{figure}

For eye-gazing interactivity, the instructor delivers a presentation or lecture
by facing the \textit{Kinect} without wearing \textsc{hdm}, as shown in
Fig.~\ref{fig:overview} (bottom).  This is a common scenario for online course
instruction, where eye contact can be established between the instructor and
student. The big screen illustrates student perception from the
\textsc{hmd} remotely. The instructor's image is rendered on the display based
on the user's view perspective. Since the imaging process strictly follows the
camera projection principle from 3D-to-2D, it delivers realistic views, which
are faithful to the physical size of the captured object. Hence, observers can
have a sensation of experiencing a face-to-face lecture. 

To enhance mutual understanding and facilitate teaching activities, our second
distinct functional module offers richer non-verbal interaction between users.
By empowering the instructor to demonstrate and manipulate objects in virtual
space, the capabilities of the system enable the instructor to capture and
sustain more student attention through concrete examples.  Concomitantly, in
response students can perform similar operations on the targeted objects in an
interactive manner. The system is responsible for incorporating the input from
each perspective (student and instructor) and synchronize the shared view on
the virtual object accordingly. We hypothesize that such interactivity is
especially useful to explain  complex concepts. For example, in an engineering
mechanics course, it is often challenging to verbally or textually describe a
sophisticated system. Students need to visually see the structure and work flow
to gain deep, profound understanding of the system.


\subsection{The Pipeline}

Fig.~\ref{fig:pipeline} describes the system's pipeline, which is comprised of
six different functional components. According to their
responsibilities, these components can be further categorized into four groups, as
indicated by different colors: blue for vision task; dark gray for graphics
task; black for virtual view rendering; and light gray for video compression
for transmission.  These components are distributed on each user's end and
provide remote communication across network.

\textbf{Vision Task:} as the input stage processing, where color and depth data
are captured from the \textit{Kinect} device; meanwhile, the
\textit{LeapMotion} controllers obtain a grayscale stereo image of the
near-infrared light spectrum, separated into the left and right cameras. These
two input streams feed into the \textsc{rgb-d} acquisition, viewpoint tracking,
and gesture detection components, respectively.  These components produce three
outcomes: the reconstructed 3D of the captured scene with textures and user'
viewpoint position, and user's corresponding pose estimation.

\textbf{Graphics Task:} In addition to the sensor's input, our system uses
pre-stored 3D models to synthesize virtual scenes, which is the main
responsibility of the 3D transformation component. For each user, the same copy
of models is stored and loaded locally to avoid heavy data transmission. During
running time, only the updated information are synchronized across the network,
such as object position and orientation changes.  This component also offers
collision detection, which is essential for realistic rendering and correct
user interactions.

\textbf{Virtual View Rendering:} the reason this component is separated from
the graphics tasks is due to its mixture of input from multiple resources. One
input is the detected local user's viewpoint position, which plays a role as
the center of a virtual camera that projects 3D data onto the 2D image plane
for dynamic perspective rendering. Another input is the pre-stored virtual
objects, including user's virtual body. In addition to the local input, remote
data also contribute to the final view synthesis. For instance, the captured
(color and depth) images of other users are transmitted through the network and
merge with local virtual objects.

\textbf{Video Compression:} to ensure realtime performance, we also need to
consider the data exchange of large visual contents across the network.  Some
video compression or tracking strategies can be developed to boost the
transmission speed.

\section{Key Component Technologies}

\subsection{Accurate 3D acquisition}

The 3D point cloud and its color texture information are obtained by the
\textit{Kinect} camera. The captured depth image has the depth value available
for each pixel. Based on the intrinsic parameters of the camera and the
measured depth, we obtain the corresponding 3D point by applying an inverse
camera projection operation. Thus, for each pixel on the color image, we can
compute the tuple \{$X, Y, Z, R, G, B$\} where $R$, $G$, $B$ are the three
color channels.

To project users into an immersive virtual environment, we need to first
extract their texture and geometry from the acquired 3D point cloud.  Depth
value for each pixel provides a useful clue allowing us to easily separate the
foreground and background. However, noise present in  the  depth  image can
significantly  impair the  quality  of  the  separation. To handle this
problem, we use our proposed two-layer pixel labeling strategy through a
probabilistic framework that incorporates background and measurement modeling
as well as available observations in the missing depth neighborhood \cite{Shen12}. The labeling procedure is formatted as a \textit{Maximum a Posteriori} (\textsc{map}) problem:

\begin{equation}
X_G^{map} \triangleq \arg\max_{x_G}(\sum_{s, t}log\psi(x_s, x_t) + \sum_slog\phi(x_s))\nonumber
\end{equation}

where, $s$ and $t$ defines the neighbor pixel indices around the target pixel.
The spatial term $\psi(x_s, x_t)$ and prior term $\phi(x_s)$ are derived from the
depth continuity and color similarities. To solve this \textit{Markov Random
Filed (\textsc{mrf})} configuration, an optimal solution can be approximated by
using \textit{Loopy Belief Propagation (\textsc{lbp})}~\cite{Freeman00}.

\subsection{View Rendering}

For dynamic virtual view rendering, a key task is to accurately track the
viewpoint in realtime. To aid this procedure, we can possibly rely on three
resources: color image, depth image, and the Oculus positioning tracker. As
each has its own limitations, it would be beneficial if we can let the
three trackers work collaboratively in a boosting manner. For the depth-based tracking, our earlier work can be adopted by approximating
the head as a sphere and treating the center as our target viewpoint
\cite{Shen13}. A complete head silhouette is obtained by inferring a curvature
curve through holes filling and outline smoothing by the
\textit{Morphological opening and closing} technique. However, the depth sensor
fails to accurately detect the target when it is out of the range [$80cm$,
$400cm$]. As a compensation, the Oculus positioning tracker and color image can
provide additional cues. An easy and na{\"{i}}ve way is to use thresholding to switch
the responsibilities between these three resources. Depth tracking is performed
when the target drops into a valid range. Otherwise, we can apply a
color-tracking algorithm, such as \textit{Camshift}~\cite{Bradski98}, or
\textit{Oculus positioning}, where the accuracy can be further enhanced in a \emph{MapReduce}-like framework \cite{wang2012scimate}~\cite{SmartReport}. 
The estimated center of the circle on the camera
plane is temporally smoothed with a \textit{Kalman filter}. This simple but
straightforward method works well for us.  However, more advanced approaches
can be employed for improved robustness and generality. For example, the
particle filter for \textsc{rgb-d} scene can be applied to solve the
6-\textsc{dof} pose tracking problem ~\cite{Choi13}, where the posterior
probability density function $p(\mathbf{X}_t|\mathbf{Z}_{1:t})$ for estimating
object trajectory at time $t$ is represented as a set of weighted particles:
$\mathcal{S}_t = \{(\mathbf{X}_t^{(1)}, \pi_t^{(1)}), ..., (\mathbf{X}_t^{(N)},
\pi_t^{(N)})\}$. The whole tracking process can also be  adapted to a multi-agent system~\cite{Luo2003}. 

\subsection{Interactivity}

The core of interactivity between users in virtual environment is the collision
detection, which imposes constraints to an object's motion by collisions with
other objects. The \textit{LeapMotion} device offers an efficient hands and
arms detection in realtime. As the estimated gesture is from egocentric
perspective (i.e., first-person view), we need first to convert the 3D
positions from each user's end to the global coordinate by using the
extrinsics. For each user $i$, according to its predefined position
$\mathbf{t}$ and orientation $\mathbf{R}$ in the virtual space, a detected 3D
point $\mathbf{p}$ from the user's perspective can be transformed to the global
coordinate as: $\mathbf{p'} = [\mathbf{R}^{(i)}, -\mathbf{R^{(i)}t^{(i)}};
\mathbf{0}, 1]\cdot \mathbf{p}$.  After the coordinate synchronization, we need
to identify any axis-alignment or overlapping between all the user's 3D points
$\mathbf{p'}$ and the virtual environments. An intuitive solution is to use
bounding boxes through dimension reduction, such as \textit{Binary Space
Partition (\textsc{bsp})}method~\cite{Thibault87}.  For our prototype
implementation, we wrap up each object by the axis-aligned bounding boxes as
bounding volumes \cite{Cohen95}. The method assumes that if two bodies collide in three
dimensional space, their orthogonal projection onto the $xy$, $yz$, and
$xz$-planes and $x$, $y$, and $z$-axes must overlap. The \textit{Sweep and Prune}
algorithm is adopted to reduce the number of pairwise collision tests by
eliminating polytope pairs~\cite{Cohen95}. To accommodate more complex models
in large environments, parallelization-based strategies can be employed for
further optimization, such as the \textit{Potentially Colliding Set
(\textsc{pcs})} approach, which can be conveniently adapted for \textsc{gpu}
processing~\cite{Govindaraju03}.

\subsection{Network Transmission}

For realtime performance, we adopt a client-server, distributed architecture to
address remote environment and high computational complexity. As depicted in
Fig.~\ref{fig:pipeline}, each client is responsible for an \textsc{rgb-d}
camera, a motion sensor, and local 3D models. To lighten the transmission
load, the tasks of 3D point cloud processing, viewpoint estimate, collision
detection, and rendering are all carried out at the client level. Each client
sends out a pair of \textsc{rgb} and depth images, detected view point (a 3D point
vector), body tracking estimation (multiple 3D points for the detected body
joints), and collision detection results to the server. The server then
combines all the received images, refines the estimations, and distributes
the data back to the clients. As most of the heavy transmission load
concentrates on the \textsc{rgb-d} image pairs, we applied our proposed
\textit{CamShift}-based algorithm to transmit a compressed version of images on
the network~\cite{Wang15}.

\section{The Performance }



\begin{figure}[!htb]
\centerline{\epsfig{figure=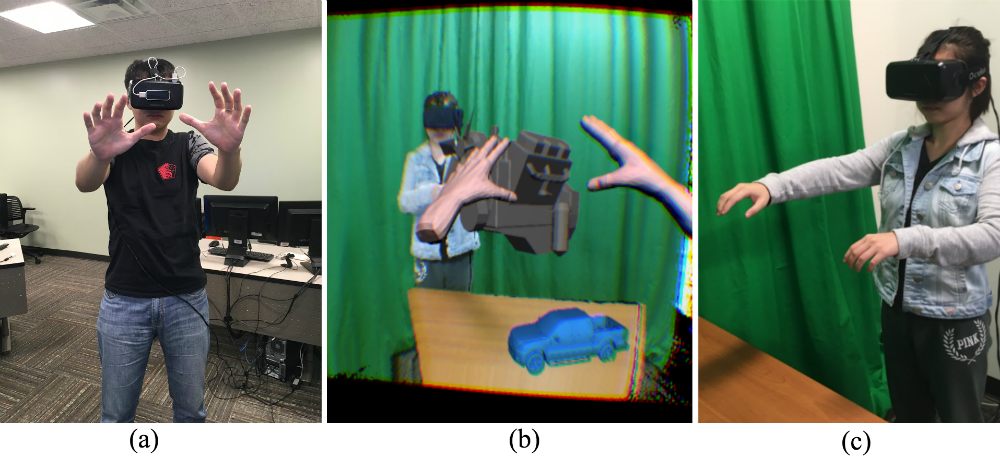, width=9cm}}
\caption{Interaction between users.}
\label{fig:interaction}
\end{figure}

\subsection{Quality Demonstrations}

\textbf{Scenario 1:} This scenario shows how users coordinate and share
resources with each other. Fig.~\ref{fig:interaction}(a) and (c) demonstrate a
scenario in which two people collaborate remotely. Fig.~\ref{fig:interaction} (b)
shows what the user (a) can observe from the \textsc{hmd}: user's hands
(presented by the virtual 3D model), shared virtual objects, and the
collaborator's image captured remotely. The example shows the user raising up a
3D model for the collaborator to catch.  This process is synchronized and
delivered to the \textsc{hmd} of both users, which allow them to perceive such
interactivity virtually.

\begin{figure}[!htb]
\centerline{\epsfig{figure=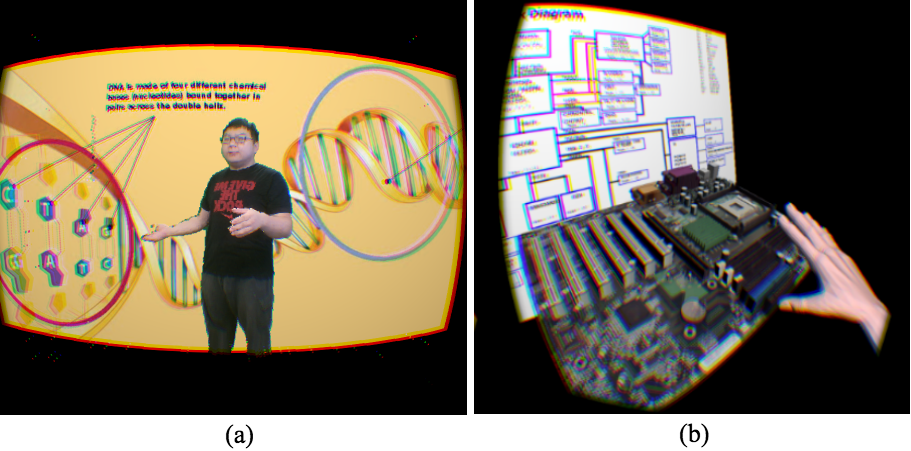, width=9cm}}
\caption{Face-to-face lecture and online practice.}
\label{fig:lecturing}
\end{figure}

\textbf{Scenario 2:} In addition to interactivity, advantages for lecturing can
be identified.  Consider online tutoring applications.
Fig.~\ref{fig:lecturing}(a) demonstrates how a learner can perceive the
lecturer from the \textsc{hmd}. A true face-to-face sensation is created by
offering life-size views and flexible distance adjustment. Based on the topic
of the lecture, different immersive virtual environments can be rendered, which
can help the lecturer provide more vivid, clear explanation.

\textbf{Scenario 3:} Traditional online lectures often suffer from a one-way
mode of communication, where each student acts as a passive observer.
Alternatively, our system allows students to actively participate in the
learning process.  Fig.~\ref{fig:lecturing}(b) shows student exercises in a
motherboard assembling course. The student can virtually practice hardware
installation while reading instructions.

\subsection{Quantity Evaluation}

\begin{figure}[!htb]
\centerline{\epsfig{figure=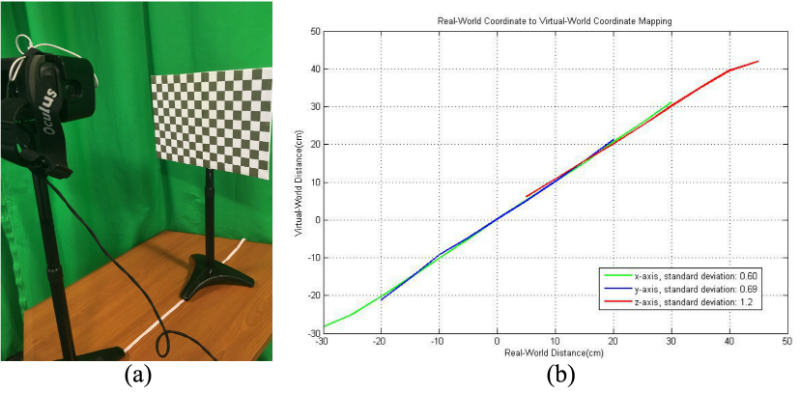, width=9cm}}
\caption{Accuracy measurement between the virtual and physical worlds.}
\label{fig:accuracy}
\end{figure}

\textbf{Accuracy: } To verify the accuracy of interactivity, we compare the
estimated body gestures in the virtual space with the ground truth, which is
based on the physical measurement of the distances between the \textsc{hmd} and
the checkerboard (see Fig~\ref{fig:accuracy}(a)). We let users place their
hands at different positions against the checkerboard with their $x, y, z$
coordinates within the ranges: $x\in[-30cm, 30cm]$, $y\in[-20cm, 20cm]$, and
$z\in[5cm, 40cm]$. Fig~\ref{fig:accuracy}(b) shows the contrast between the
gesture estimation in the virtual world and physical measurement in the real
world along the $x, y, z$-axes. The standard deviation of their offset is
managed to be less than $1.2 cm$ among all the directions. 

\textbf{Time performance:} The preliminary prototype is implemented using C++,
OpenCV, and OpenGL library, running on each client machine with an Intel Core
i7-4790 (8M Cache, 4.0), 16GB Dual Channel \textsc{ddr}3 1600MHz \textsc{ram}.
The computational performance for each component which is measured in per-frame
processing speed, is given in Table~\ref{tab:results}.  We tested different 3D
model meshes with the average time cost at approximately $25 ms$ for processing
$3.3M$ polygons. Without any parallelization, our method can process $424\times
512$ video at $14fps$, which closely approaches an interactive speed.

\begin{table}
\centering
\caption{ .}
  \begin{tabular}{|l||l|l|}
    \hline
     & \textbf{Processing Time} & \textbf{Rendering Time} \\ \hline \hline
    3D Object & 25 ms & 9 ms \\ \hline
    \textsc{rgb-d} Processing & 23 ms & $<$ 1 ms \\ \hline
    Collision Detection & $<$ 1 ms & $<$ 1 ms \\ \hline
    Network Transmission & 5 ms &  -- \\ \hline
    \textbf{Total Time} &  \multicolumn{2}{|l|}{$\mathbf{\approx 73 ms}$ (framerate $\mathbf{\approx 14 fps}$)} \\
    \hline
  \end{tabular}
\label{tab:results}
\end{table}

\section{Educational Applications}

Immersive learning experiences such as those which our system helps support
have the potential to alter the way educators think about and deliver primary
and higher education, and the methods by which students engaging in
experiential learning~\cite{hufpost}.

Virtual environments such the system we describe, which support the
synchronous, physical manipulation of objects by participants in a
collaborative, mixed-initiative motif~\cite{mii}, both blur the boundaries
between teacher and student, and foster enhanced student-student collaboration.
While the educational applications of such collaborative, simultaneous object
manipulation are diverse and countless from teaching children to write to
advanced military training exercises, we focus on two broad and progressive
application categories within education here and make some cursory remarks.

\begin{itemize}

\item \textit{Developing motor skills in kindergarten children}: An important,
early educational application of our system is the development of motor skills
among pre-adolescent children, where mutual manipulation of objects is an
integral part of the learning process.  Such an application can evolve into,
for example, teaching young children penmanship in kindergarten (i.e., a
teacher can help the pupil hold a pencil in her hand and demonstrate proper
technique) or support for robotic-oriented ways to teach computer programming
to children as made possible through the \textit{\textsc{kibo}} robot
kit\footnote{http://tkroboticsnetwork.ning.com/kibo}.

\item \textit{\textsc{Stem} applications across a range of high school
subjects, especially using virtual modeling and simulation technologies}: Our
work also can support teachers and students helping each other explore and
manipulate objects in the study of physical, astronomical, biological,
chemical, and computational phenomena.  These objects can exist in real-world,
physical models or virtual simulations.  For instance, students can
collaboratively and simultaneously manipulate 3D object models of molecular
structures, chemical bonds, solar systems, or even multi-dimensional hypothesis
spaces in the study of machine learning.

\end{itemize}

A recent article in the Huffington Post showcases seven cyberlearning
technologies including `tools for real-time collaboration,' an area in which
our work lies, funded by \textsc{nsf} that are transforming education~\cite{hufpost}.
Our work is focused on collaborative manipulation of objects in virtual
environments and, thus, can complement and support many of these efforts,
especially in the two areas of education briefly mentioned above.  Chris
Hoadley, the program officer at \textsc{nsf} who leads the Cyberlearning program, has
said: ```I believe it's only by advancing technology design and learning
research together that we'll be able to imagine the future of
learning'''~\cite{hufpost}.

\section{Conclusion}

We have presented the design and realization of an immersive telepresence
system by employing \textsc{rgb-d} systems, motion sensors, head-mounted
displays, and networking setup. We demonstrated the implementation and
optimization strategies of each component and integrated them systematically.
The system can be run in realtime, which is indeed a proof of concept for
practical applications. We addressed its potential benefits for educational
applications, particularly in cyberlearning. One of our future plans is to
conduct an extensive evaluation of users' experiences using the proposed system
for a variety of learning tasks. From a technical perspective, to improve the
current prototype, related key technologies will be explored to deal with more
complex environments and an increased number of users.


\section*{Acknowledgment}

This work was supported in part through a seed grant from the University of
Dayton Research Council.  The authors thank students Xiaotian Yang, Jiaqian
Zhang, Wangyufan Gong, and Xinhe Peng for assistance during experiments.

\end{document}